\documentclass{article}

\usepackage{microtype}
\usepackage{graphicx}
\usepackage{caption}
\usepackage{subcaption}
\usepackage{booktabs} 

\usepackage{hyperref}

\usepackage[preprint]{icml2026}

\usepackage{amsmath}
\usepackage{amssymb}
\usepackage{mathtools}
\usepackage{amsthm}
\usepackage{tcolorbox}

\usepackage[capitalize,noabbrev]{cleveref}

\theoremstyle{plain}

\theoremstyle{definition}

\theoremstyle{remark}

\usepackage[textsize=tiny]{todonotes}

\icmltitlerunning{}

\begin{document}

\twocolumn[
  \icmltitle{A\textsuperscript{2}-LLM: An End-to-end Conversational Audio Avatar Large Language Model}



  \icmlsetsymbol{equal}{*}

  \begin{icmlauthorlist}
    \icmlauthor{Xiaolin Hu}{equal,BUPT,LAB}
    \icmlauthor{Hang Yuan}{equal,ECNU,LAB}
    \icmlauthor{Xinzhu Sang}{BUPT}
    \icmlauthor{Binbin Yan}{BUPT}
    \icmlauthor{Zhou Yu}{ECNU}
    \icmlauthor{Cong Huang}{LAB}
    \icmlauthor{Kai Chen}{LAB}
  \end{icmlauthorlist}

  \icmlaffiliation{BUPT}{State Key Laboratory of Information Photonics and Optical Communications, Beijing University of Posts and Telecommunications, Beijing, China}
  \icmlaffiliation{LAB}{Zhongguancun Institute of Artificial Intelligence, Beijing, China}
  \icmlaffiliation{ECNU}{School of Finance and Statistics, East China Normal University, Shanghai, China}

  \icmlcorrespondingauthor{Xinzhu Sang}{xzsang@bupt.edu.cn}
  \icmlcorrespondingauthor{Kai Chen}{kaichen@zgci.ac.cn}

  \icmlkeywords{Multimodal Large Language Models; Human-Computer Interaction; Audio-Motion Generation}
    
  \vskip 0.3in
]



\printAffiliationsAndNotice{}  

\begin{abstract}
  Developing expressive and responsive conversational digital humans is a cornerstone of next-generation human-computer interaction. While large language models (LLMs) have significantly enhanced dialogue capabilities, most current systems still rely on cascaded architectures that connect independent modules. These pipelines are often plagued by accumulated errors, high latency, and poor real-time performance. Lacking access to the underlying conversational context, these pipelines inherently prioritize rigid lip-sync over emotional depth. To address these challenges, we propose A\textsuperscript{2}-LLM, an end-to-end conversational audio avatar large language model that jointly reasons about language, audio prosody, and 3D facial motion within a unified framework. To facilitate training, we introduce FLAME-QA, a high-quality multimodal dataset designed to align semantic intent with expressive facial dynamics within a QA format. By leveraging deep semantic understanding, A\textsuperscript{2}-LLM generates emotionally rich facial movements beyond simple lip-synchronization. Experimental results demonstrate that our system achieves superior emotional expressiveness while maintaining real-time efficiency (500 ms latency, 0.7 RTF). The code and dataset will be available at \href{https://github.com/Hooyoung-for-AI/A2-LLM.git}{the project repository}.
\end{abstract}

\section{Introduction}
\label{sec:intro}

Large language models (LLMs) have rapidly evolved from text-only systems to multimodal agents capable of processing and generating speech, enabling more natural forms of Human--Computer Interaction (HCI). In particular, recent Large Audio-Language Models (LALMs) have demonstrated strong performance in speech understanding and generation, surpassing conventional cascaded approaches and enabling direct, real-time spoken interaction.

Among existing systems, GPT-4o~\cite{openai_gpt4o_system_card} represents the first end-to-end audio-interactive large model, directly taking audio input to audio output without relying on intermediate text representations, thereby preserving paralinguistic cues and global conversational coherence. Following GPT-4o, a series of open-source LALMs have emerged~\cite{fang2024llama, zeng2024glm, fang2025llama, ding2025kimi, wu2025step, huang2025step, li2025baichuan}, among which Step-Audio2 demonstrates particularly strong emotional speech understanding and generation. Despite these advances, current LALMs remain limited to audio-only interaction, which is insufficient for many  HCI scenarios~\cite{cassell2001embodied}. In natural face-to-face communication, speech and facial expressions are tightly coupled, and humans rely on both auditory and visual cues to infer intent and emotion~\cite{mcgurk1976hearing, yehia1998quantitative}.

To complement audio interaction with visual feedback, recent video generation models~\cite{xu2024vasa, xu2024hallo} have achieved impressive photorealism. However, their reliance on 2D pixel generation lacks underlying 3D geometry, precluding deployment in immersive VR/XR and naked-eye 3D displays which demand stereoscopic 6DoF consistency.

Consequently, 3D-native facial animation remains the standard for immersive human-computer interaction. The prevailing paradigm involves mapping speech signals directly to the parameters of FLAME~\cite{li2017learning}, a widely adopted 3D Morphable Model (3DMM) that efficiently disentangles facial geometry into shape, expression, and pose attributes. Crucially, integrating these parametric controls with advanced neural rendering pipelines~\cite{grassal2022neural, kerbl3Dgaussians, qian2024gaussianavatars} reconciles high-fidelity synthesis with geometric consistency, satisfying the requirements of such immersive environments.

To enable face-to-face interaction, systems typically adopt a cascaded architecture (ASR $\rightarrow$ LLM $\rightarrow$ TTS $\rightarrow$ Animation)~\cite{huang2024audiogpt, zhang2023speechgpt}. While modular, this pipeline inherently suffers from high latency, accumulated errors, and a fundamental \textit{Semantic-Emotion Gap}. By compressing LLM reasoning into audio waveforms, conversational context is lost, often leading to mismatched expressions—for instance, articulating laughter (``haha'') with mechanically synchronized lips but a stiff, emotionless upper face. Recent attempts mitigate this by either injecting explicit emotion conditions (e.g., DiffPoseTalk~\cite{sun2024diffposetalk}, ARTalk~\cite{chu2025artalk}) or patching extracted semantic features onto decoupled backbones (e.g., Wav2Sem~\cite{li2025wav2sem}). Yet, both strategies remain constrained by their decoupled nature: focusing primarily on lip-shape reconstruction, they fail to achieve genuine, conversational-level emotional understanding.

In this work, we address these challenges by proposing \textbf{A\textsuperscript{2}-LLM}, an end-to-end conversational audio avatar large language model. Departing from decoupled paradigms, our framework jointly reasons about language, audio prosody, and 3D facial motion within a unified modality space. Specifically, we discretize facial dynamics into hierarchical tokens via an RVQ-VAE \cite{zeghidour2021soundstream} and bridge them to the language backbone using a specialized Motion Connector. By jointly decoding synchronized audio and facial tokens, A\textsuperscript{2}-LLM eliminates the \textit{Semantic-Emotion Gap} at the source. This holistic approach not only minimizes cascading errors and system latency but also ensures that facial behaviors—from lip sync to upper-face expressions—are deeply grounded in the conversational context rather than merely inferred from acoustic surfaces.

To facilitate training, we introduce \textbf{FLAME-QA}. Unlike conventional datasets restricted to unstructured Audio-Visual (AV) pairs \cite{nagrani2017voxceleb, afouras2018lrs3, zhang2021flow}, our dataset is organized as Multimodal QA triplets. Specifically, each sample pairs an input question with a multimodal response, which encompasses both synchronized speech and facial dynamics. This inclusion of an antecedent question imposes a semantic constraint absent in prior work: it compels the model to generate facial behaviors that are not merely synchronized with the acoustic signal, but are logically conditioned on the dialogue context.

In summary, our contributions are:

\textbf{FLAME-QA Dataset:} We introduce the multimodal dataset structured as QA triplets. Unlike unstructured audio-visual pairs, this format provides strictly aligned semantic supervision, ensuring facial dynamics are logically conditioned on the dialogue context.
    
\textbf{A\textsuperscript{2}-LLM Framework:} We propose a unified framework that jointly decodes audio and facial tokens. This end-to-end approach eliminates the \textit{Semantic-Emotion Gap}, enabling highly responsive ($\sim$500 ms) and emotionally grounded conversational interaction.

\textbf{Generative Facial Capability:} We systematically validate the latent alignment between large-scale semantic understanding and expressive 3D motion. By leveraging an RVQ-VAE and Motion Connector, we demonstrate that fine-grained, context-aware facial dynamics can be effectively grounded in the model's native semantic processing.

\section{Related Works}
\label{sec:rel}

\textbf{Evolution of End-to-End LALMs.}
The paradigm of spoken conversational agents is shifting from high-latency modular cascades to unified Large Audio-Language Models (LALMs) \cite{huck2025spoken}. Recent innovations like JoyVoice \cite{yu2025joyvoice}, EMOVA \cite{chen2025emova}, and Step-Audio \cite{wu2025step} integrate speech processing directly into the LLM backbone, achieving rich emotional prosody and handling complex multi-speaker contexts. However, despite the auditory expressiveness of models like GPT-4o~\cite{openai_gpt4o_system_card}, they remain "disembodied." They excel at generating acoustic signals but lack the native capability to control a 3D morphological embodiment, creating a significant void in visual face-to-face interaction where auditory and visual cues are inherently coupled.

\textbf{Generative Avatars and Emotional Alignment.}
To visualize speech, research has evolved into two main streams: 2D video synthesis and 3D facial animation. While 2D diffusion models \cite{low2025talkingmachines} achieve photorealism, they lack the geometric consistency required for immersive VR/XR. Conversely, 3D-native methods typically rely on audio-driven regression. As highlighted by HCI research \cite{zhang2025ai, llanes2024developing}, such decoupled approaches often suffer from a \textit{Semantic-Emotion Gap}—generating generic lip movements that fail to reflect the conversation's emotional context (e.g., a sad tone with a neutral face), which significantly increases user cognitive load. Unlike purely audio-reactive agents \cite{park2024let}, A\textsuperscript{2}-LLM utilizes discrete motion tokenization to treat facial dynamics as a language. By training on our proposed FLAME-QA dataset, we ensure that generated expressions are not merely synchronized with audio, but are logically grounded in the semantic intent of the dialogue.

\begin{figure*}[t]
    \centering
    \includegraphics[width=1.0\linewidth]{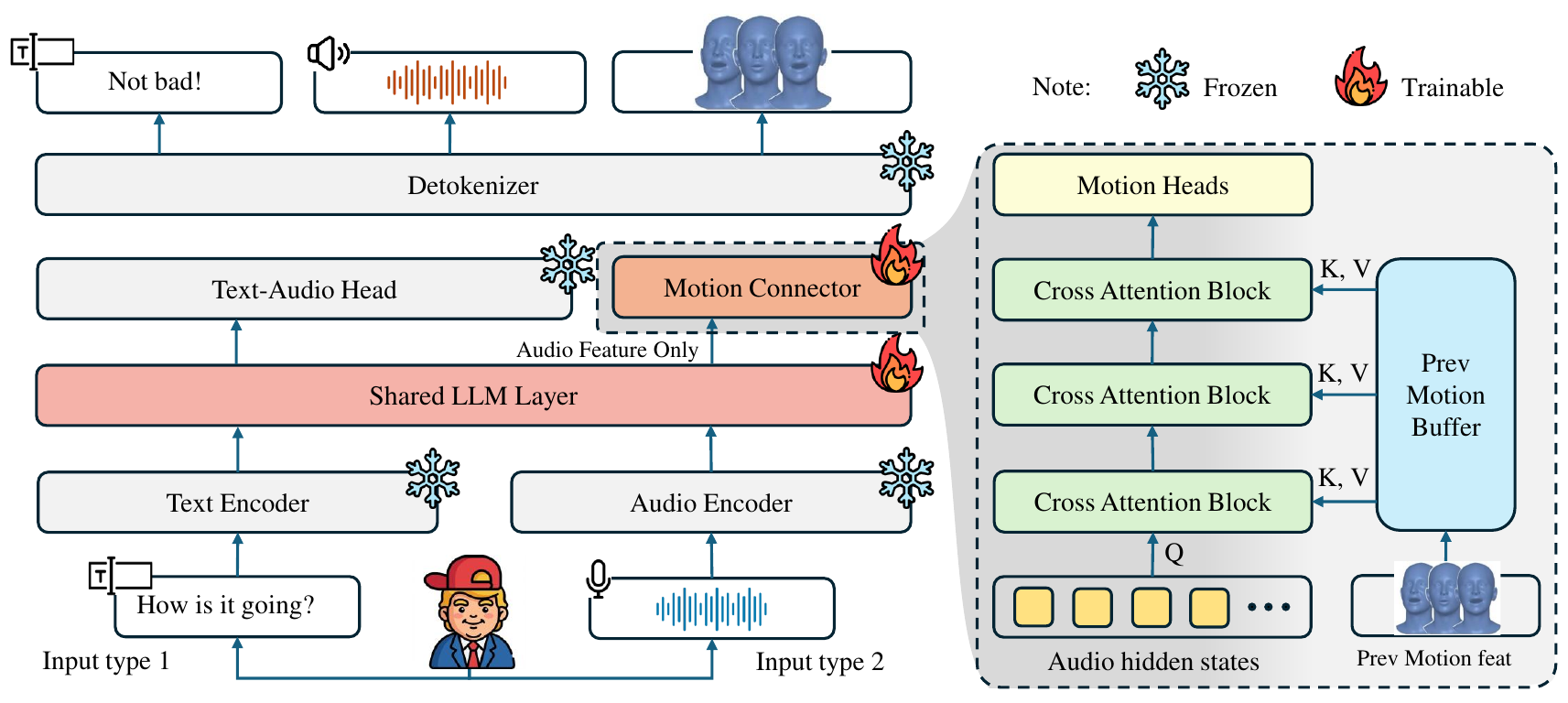}
    \caption{\textbf{Schematic of A\textsuperscript{2}-LLM.}
\textbf{Left:} The unified framework where the Shared LLM Layer and Motion Connector are jointly optimized to generate synchronized speech and facial dynamics.
\textbf{Right:} The Motion Connector predicts hierarchical tokens via Motion Heads. It adopts a segment-wise autoregressive design, utilizing audio hidden states (serving as Queries) to attend to the buffered motion history (Keys and Values) for temporal continuity.}
    \label{fig:framework}
\end{figure*}

\section{Preliminaries}
\label{sec:prelim}

\subsection{End-to-End Audio-Language Models}
\label{sec:prelim:lalm}
To bypass the latency and error propagation inherent in cascaded systems, we build upon the architecture of Step-Audio-2-mini~\cite{wu2025step}, an end-to-end LALM. Initialized from Qwen2.5-7B \cite{xu2025qwen2} and utilizing the encoder from Qwen2-Audio, this model is designed to perceive and generate speech directly without intermediate textual bottlenecks. The architecture integrates an audio encoder, adaptor, LLM decoder, and a flow-matching detokenizer. Crucially, the decoder autoregressively generates a unified, interleaved sequence of discrete text and audio tokens. This synchronization ensures that the decoder's hidden states intrinsically capture both semantic reasoning and acoustic prosody, yielding a rich, temporally aligned feature space essential for driving expressive facial motion.

\subsection{FLAME 3D Facial Representation}
\label{sec:prelim:flame}
To provide the conversational agent with a geometrically consistent embodiment, we utilize the FLAME 3D Morphable Model~\cite{li2017learning}. FLAME mathematically disentangles facial geometry into low-dimensional parameters controlling identity, pose, and expression. Formally, a 3D face mesh $\mathbf{V} \in \mathbb{R}^{5023 \times 3}$ is reconstructed via a differentiable function $M(\beta, \psi, \theta)$:
\begin{equation}
    \mathbf{V} = M(\beta, \psi, \theta)
\end{equation}
where $\beta \in \mathbb{R}^{|\beta|}$ denotes \textit{shape} parameters (identity), $\psi \in \mathbb{R}^{|\psi|}$ denotes \textit{expression} parameters, and $\theta \in \mathbb{R}^{|\theta|}$ represents \textit{pose} parameters (neck, jaw, and global rotation).

In this work, we focus on synthesizing dynamic conversational behaviors. We fix the identity shape $\beta$ and model the time-variant state vector $\mathbf{x}_t \in \mathbb{R}^{D}$ at each frame $t$. Specifically, our configuration utilizes $50$ expression parameters, $3$ jaw pose parameters, $3$ global pose parameters, and $2$ eyelid parameters, resulting in a compact motion space of dimension $D=58$.

\subsection{Problem Formulation}
\label{sec:problem}

We formulate the task as \textit{Joint Multimodal Generation}. Given dialogue context $\mathbf{C}$ and user audio $\mathbf{A}^{in}$, the goal is to synthesize a response $\mathbf{R}$ consisting of three synchronized modalities: text $\mathbf{Y}^t$, speech audio $\mathbf{Y}^a$, and 3D facial motion $\mathbf{Y}^m$. 
Specifically, we represent the facial motion modality as a continuous sequence of FLAME parameters $\mathbf{M} = (\mathbf{x}_1, \dots, \mathbf{x}_T) \in \mathbb{R}^{T \times D}$, where $\mathbf{x}_t$ denotes the frame-wise state vector defined in Sec.~\ref{sec:prelim:flame}.

The generation process is decomposed into two coupled stages. First, the LALM backbone autoregressively generates interleaved text and audio tokens. To ensure tight audio-visual synchronization without inflating the context length, we adopt an \textit{Audio-Anchored} strategy for motion. Specifically, we extract the sequence of hidden states $\mathbf{H}$ emerging from the LLM during the audio generation process and map them to facial motion tokens via a learnable connector $\mathcal{F}_{\theta}$:

\begin{equation}
    \mathbf{Y}^m_t = \mathcal{F}_{\theta}\big(\mathbf{H}_t, \mathbf{Y}^m_{<t}\big)
\end{equation}

This formulation ensures $\mathbf{Y}^m$ is grounded in both the semantic reasoning of $\mathbf{Y}^t$ and the acoustic prosody of $\mathbf{Y}^a$ captured within $\mathbf{H}$.

To support this objective, we require data formatted as Multimodal Question-Response pairs. Formally, we define a training sample as a tuple $(\mathbf{Q}, \mathbf{R})$, where the response $\mathbf{R}$ is decomposed into aligned modalities: $(\mathbf{Q}, \mathbf{R}_{text}, \mathbf{R}_{audio}, \mathbf{R}_{visual})$.
The lack of such instruction-following audio-visual data motivates our construction of the \textbf{FLAME-QA} dataset (Sec.~\ref{sec:dataset}).

\section{FLAME-QA Dataset}
\label{sec:dataset}

Existing datasets like TFHP~\cite{sun2024diffposetalk} and HDTF~\cite{zhang2021flow} provide only raw audio-visual pairs, lacking the instruction-following structure required for our problem formulation \cite{wei2021finetuned, wang2023self, liu2023visual, ouyang2022training}. To address this, we construct FLAME-QA, a large-scale multimodal dataset comprising $\sim$100k high-quality samples, specifically filtered and formatted for end-to-end avatar training.

\noindent\textbf{Data Curation Pipeline.}
Starting from a pool of 800k raw clips from VoxCeleb~\cite{Nagrani19}, we implement a rigorous dual-stream processing pipeline. In the visual stream, we utilize SMIRK~\cite{retsinas20243d} to extract high-fidelity FLAME parameters ($\beta, \psi, \theta$) from each frame as ground-truth motion targets. Simultaneously, the acoustic-semantic stream employs Whisper~\cite{radford2022whisper} for ASR transcription alongside an auxiliary module to label affective states. To ensure data quality, we leverage a LLM to assess semantic coherence, filtering out segments with broken logic or unclear speech, resulting in a refined subset of 100k samples.

\noindent\textbf{Multimodal Triplet Construction.}
To satisfy the requisite $(\mathbf{Q}, \mathbf{R})$ structure defined in Sec.~\ref{sec:problem}, we must synthesize the missing antecedent instructions. For each selected transcript (serving as the ground-truth response text $\mathbf{R}_{text}$), we prompt an LLM to generate a context-aware Question $\mathbf{Q}_{text}$, which is then synthesized into speech $\mathbf{Q}_{audio}$ using IndexTTS2~\cite{zhou2025indextts2}. This process yields fully aligned triplets where the facial motion $\mathbf{R}_{visual}$ is semantically anchored to the dialogue context established by $\mathbf{Q}$.

\noindent\textbf{Task Formatting and Expression Augmentation.}
To prevent catastrophic forgetting and maintain general audio capabilities, we organize the data into four subtasks trained under a unified objective: AQAA (Audio QA), Repeat (Speech Resynthesis), TTS (Text to Speech), and ASR (Automatic Speech Recognition). Furthermore, to transcend the limited emotional range of news-centric datasets, we curate a specialized High-Dynamic Subset of $\sim$1k samples. Generated via InfiniteTalk~\cite{yang2025infinitetalk}, these samples feature extreme expressions and rich affective content (e.g., laughter, surprise), providing strong supervision for vivid emotional generation.

\section{Method}
\label{sec:met}
In this section, we introduce A\textsuperscript{2}-LLM, an end-to-end multimodal language model designed for expressive and context-aware facial motion generation. As illustrated in Figure~\ref{fig:framework}, our system departs from cascaded pipelines by jointly modeling language, audio prosody, and facial dynamics within a unified framework. We begin by defining the joint multimodal generation task and our audio-anchored modeling approach (Sec.~\ref{sec:problem}). Next, we describe the Residual Motion Tokenization (Sec.~\ref{sec:met:VQ}), which discretizes continuous facial dynamics into hierarchical tokens via a temporal RVQ-VAE. We then present the Motion Connector (Sec.~\ref{sec:met:connector}), a module that aligns these tokens with the LLM backbone using history-conditioned attention. Subsequently, we detail the model architecture and the three-stage curriculum strategy employed to balance instruction following with motion expressiveness (Sec.~\ref{sec:met:MLLM}). Finally, we define the unified training objective (Sec.~\ref{sec:met:objective}) that optimizes the system for synchronized text, audio, and motion generation.

\subsection{Residual Motion Tokenization}
\label{sec:met:VQ}

\begin{figure}[t]
    \centering
    \vspace{-8pt} 
    \includegraphics[width=0.8\linewidth]{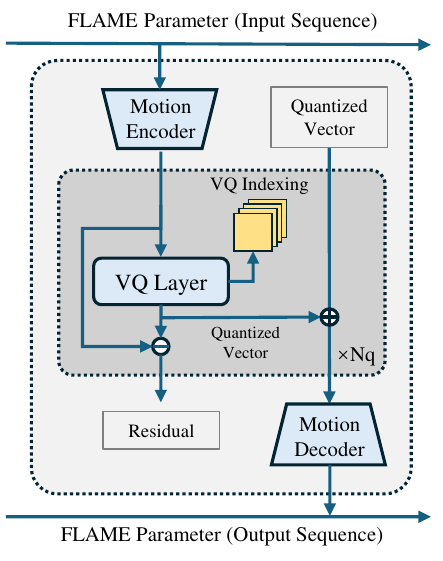}
    \vspace{-5pt}
    \caption{\textbf{Illustration of the Motion Tokenization Module.} The module encodes sequences of continuous FLAME parameters into compact hierarchical discrete tokens via Residual Vector Quantization (RVQ), enabling efficient autoregressive modeling.}
    \vspace{-5pt}
    \label{fig:vqvae}
\end{figure}

To model high-frequency facial dynamics while maintaining temporal coherence, we employ a Residual Vector Quantized VAE (RVQ-VAE)~\cite{zeghidour2021soundstream} equipped with a temporal compression mechanism. The architecture of our RVQ-VAE is shown in Figure \ref{fig:vqvae}.
Formally, given a raw motion sequence $\mathbf{M} \in \mathbb{R}^{T \times D}$ (where $T$ corresponds to the 25 Hz audio frame rate), the encoder compresses it temporally by a factor of $G$ (e.g., $G=5$). This yields a compact latent sequence $\mathbf{z} \in \mathbb{R}^{(T/G) \times d_z}$.

To capture both semantic structure and fine-grained details, we discretize $\mathbf{z}$ using Residual Vector Quantization (RVQ) with $N_q$ codebooks. The quantization process decomposes the latent vector into a sum of discrete codes:
{
\setlength{\abovedisplayskip}{8pt}
\setlength{\belowdisplayskip}{8pt}
\begin{equation}
\mathbf{q} = \sum_{j=1}^{N_q} \mathbf{e}_{k_j}, \quad \text{where } \mathbf{e}_{k_j} = \mathcal{C}_j(\mathbf{r}_{j-1})
\end{equation}
}
Here, $\mathcal{C}_j$ denotes the $j$-th codebook, and the residual is recursively defined as $\mathbf{r}_j = \mathbf{r}_{j-1} - \mathbf{e}_{k_j}$.

The overall training objective combines reconstruction fidelity with codebook commitment:
{
\setlength{\abovedisplayskip}{10pt}
\setlength{\belowdisplayskip}{10pt}
\begin{equation}
\mathcal{L}_{\text{VQVAE}} = \mathcal{L}_{\text{rec}} + \|\operatorname{sg}[\mathbf{z}] - \mathbf{q}\|_2^2 + \gamma \|\mathbf{z} - \operatorname{sg}[\mathbf{q}]\|_2^2
\end{equation}
}
where $\operatorname{sg}[\cdot]$ denotes the stop-gradient operator. To ensure geometrically accurate and temporally stable animation, we do not rely solely on parameter regression. Following ARTalk~\cite{chu2025artalk}, we formulate $\mathcal{L}_{\text{rec}}$ by explicitly incorporating vertex-level supervision (targeting lips and face regions) and temporal derivative constraints (velocity and acceleration), thereby enforcing precise lip-synchronization and smooth motion transitions.

\subsection{Motion Connector with Hierarchical Context}
\label{sec:met:connector}

The Motion Connector bridges the modality gap between the audio-centric LLM and the discrete motion space.
As formulated in the backbone architecture, the LLM decoder generates audio tokens at a high frequency (25 Hz). Consequently, for the current processing window, we obtain a sequence of audio-aligned hidden states $\mathbf{H} \in \mathbb{R}^{T \times d}$.

\textbf{Feature Downsampling and Alignment.}
Since the target RVQ tokens operate in a compressed temporal space (as defined in Sec.~\ref{sec:met:VQ}), a direct frame-to-frame mapping is ill-posed. We therefore perform feature-level downsampling via a 1D convolution layer followed by a linear projection. This maps the input $\mathbf{H}$ to the compressed motion resolution $\mathbf{H}' \in \mathbb{R}^{(T/G) \times d}$, strictly aligning the semantic features with the target motion token rate.

\textbf{History-Conditioned Attention.}
To ensure smooth transitions and temporal continuity across segments, we adopt a Segment-wise Autoregressive mechanism via a Cross-Attention Transformer.
Instead of generating motion in isolation, we explicitly condition the current prediction on the motion trajectory of the preceding temporal segment.
Specifically, the discrete motion tokens from the previous segment are projected into dense embeddings to serve as the context memory. In the Transformer layers, the current downsampled audio features $\mathbf{H}'$ act as Queries, while the history motion embeddings serve as Keys and Values. This design allows the model to retrieve historical motion states to maintain coherence while synthesizing new dynamics driven by the current audio prosody.

\textbf{Multi-Scale Hierarchical Prediction.}
To capture the full spectrum of facial dynamics—from semantic jaw movements to fine-grained micro-expressions—our connector is designed to predict the complete set of hierarchical RVQ indices.
We utilize $N_q$ layer-specific prediction heads, where each head is responsible for modeling the probability distribution of the $k$-th quantization layer.
The system is optimized via a multi-task learning objective, aggregating the Cross-Entropy loss across all hierarchical levels:
{
\setlength{\abovedisplayskip}{8pt}
\setlength{\belowdisplayskip}{8pt}
\begin{equation}
    \mathcal{L}_{motion} = \sum_{k=1}^{N_q} \mathcal{L}_{CE}(\mathbf{q}^{(k)}, \hat{\mathbf{q}}^{(k)})
    \label{eq:motion_loss}
\end{equation}
}
where $\mathbf{q}^{(k)}$ represents the ground-truth token indices. This formulation encourages the model to jointly learn coarse semantic structures and subtle high-frequency details within a unified context.

\subsection{Multimodal Large Language Model and Training Strategy}
\label{sec:met:MLLM}

A\textsuperscript{2}-LLM is built upon the Step-Audio-2-mini \cite{wu2025step} backbone. To endow the model with expressive facial motion capabilities while preserving its linguistic intelligence, we employ Low-Rank Adaptation (LoRA) \cite{hu2022lora} with a proposed three-stage curriculum designed to address the stability-convergence trade-offs in multimodal joint training.

\textbf{Stage 1: Motion Connector Pre-training.} Our initial objective is the convergence of the Motion Connector. Experiments indicate that a uniform learning rate leads to gradient collapse due to the high variance of the initialized Connector interacting with the pre-trained LLM. We thus adopt a differential learning rate strategy (e.g., LoRA: $1e^{-4}$, Connector: $1e^{-5}$). However, this introduces a convergence rate discrepancy: the fast-learning LoRA module rapidly overfits and suffers catastrophic forgetting, while the slow-learning Connector requires more steps to capture fine-grained dynamics. This creates a dilemma where stopping early yields jittery motion, while prolonged training degrades linguistic capability.

\textbf{Stage 2: Joint Alignment with LoRA Reset.} To resolve this, we implement a LoRA Reset strategy. Upon Connector convergence in Stage 1, we discard the overfitted LoRA weights while retaining the converged Motion Connector. We then re-initialize LoRA and perform a second round of joint fine-tuning using the same differential learning rates.In this phase, the pre-trained Connector serves as a stable semantic anchor. The fresh LoRA module aligns effectively with the motion modality in significantly fewer steps, thereby avoiding the prolonged training that leads to catastrophic forgetting.

\textbf{Stage 3: Affective Instruction Tuning.} Finally, to imbue the avatar with vivid emotional traits, we continue to train the model on our curated subset of $\sim$1k high-freedom emotional QA samples. This phase acts as style injection, enabling the system to generate nuanced, empathetic facial expressions that fully reflect the dialogue's emotional context.

\subsection{Training Objective}
\label{sec:met:objective}

We optimize the A\textsuperscript{2}-LLM framework using a standard cross-entropy loss. Consistent with the backbone's architecture, we treat text and audio tokens as a unified interleaved sequence rather than separate modalities. Consequently, the total training loss is formulated as:
{
\setlength{\abovedisplayskip}{8pt}
\setlength{\belowdisplayskip}{8pt}
\begin{equation}
    \mathcal{L} = \mathcal{L}_{\text{seq}} + \lambda \mathcal{L}_{\text{motion}}
\end{equation}
}
where $\mathcal{L}_{\text{seq}}$ denotes the autoregressive next-token prediction loss for the joint text-audio sequence, and $\mathcal{L}_{\text{motion}}$ represents the hierarchical cross-entropy loss for the predicted motion tokens (as defined in Eq.~\ref{eq:motion_loss}). The hyperparameter $\lambda$ balances the gradient magnitude between the main generative backbone and the motion connector, ensuring that semantic coherence, acoustic prosody, and facial dynamics are optimized synchronously.

\section{Experiments}
In this section, we systematically evaluated the model's performance, including its real-time capabilities, language abilities and facial expression synthesis capabilities. Finally, we conducted ablation experiments with phased training to verify the effectiveness of the training strategy.

\subsection{Real-time Behavior}
\label{sec:exp:realtime}

To validate A\textsuperscript{2}-LLM's capability as a responsive digital human, we evaluate its inference latency and processing efficiency using three key metrics. \textbf{Time To First Token (TTFT)} measures the latency from user input to the first generated audio token. \textbf{Time To First Action (TTFA)} represents the total delay before the first facial motion is generated. \textbf{Real-Time Factor (RTF)} is defined as the ratio of generation time to the duration of the synthesized content, quantifying throughput efficiency.
\begin{table}[h]
    \centering
    \caption{Real-time performance comparison. A\textsuperscript{2}-LLM (Optimized) utilizes graph compilation to minimize kernel launch overheads. Cascaded system results are preliminary estimates averaged over varying response lengths.}
    \label{tab:latency}
    \resizebox{\linewidth}{!}{
    \begin{tabular}{l|ccc}
        \toprule
        \textbf{Method} & \textbf{TTFT} (ms) & \textbf{TTFA} (ms) & \textbf{RTF} \\
        \midrule
        Cascaded & - & 13,730 & $> 1.0$ \\
        Cascaded (Streaming) & - & 3,232 & $> 1.0$ \\
        \midrule
        A\textsuperscript{2}-LLM (Base) & 47.24 & 850.15 & 1.005x \\
        \textbf{A\textsuperscript{2}-LLM (Optimized)} & 50.71 & \textbf{535.53} & \textbf{0.703x} \\
        \bottomrule
    \end{tabular}
    }
\end{table}
\textbf{Performance Analysis.} As shown in Table~\ref{tab:latency}, leveraging static graph compilation significantly reduces overhead. The optimized A\textsuperscript{2}-LLM achieves a TTFA of 535 ms and an RTF of 0.7x. This confirms that our system generates facial dynamics faster than real-time playback, ensuring fluid, non-blocking interaction.

\textbf{Comparison with Baselines.} We benchmark against standard cascaded systems. While batch processing incurs prohibitive latency ($>$13s) and even optimized streaming pipelines suffer from accumulated modular delays ($\sim$3.2s), A\textsuperscript{2}-LLM's unified architecture eliminates these bottlenecks. By jointly decoding audio and motion, our model reduces latency by an order of magnitude ($\sim$0.5s TTFA), effectively bridging the gap between processing and perception.

\subsection{Language Capabilities}
\label{sec:exp:language}

To strictly evaluate whether the integration of facial motion tokens compromises the model's linguistic intelligence, we conducted a comprehensive evaluation using the \textit{OpenVoiceBench}. Proposed by Baichuan-Audio \cite{li2025baichuan}, this benchmark is specifically designed to assess the capabilities of audio-native models across diverse domains, including logical reasoning, knowledge retrieval, and instruction following.

\textbf{Baselines.}
We benchmark A\textsuperscript{2}-LLM, which comprises an 8B LLM backbone, a 0.16B LoRA adapter, and a 0.12B motion connector, against leading end-to-end audio models: GLM-4-Voice (9B) \cite{zeng2024glm}, Baichuan-Audio (7B) \cite{li2025baichuan}, StepAudio-Chat (132B) \cite{huang2025step}, and Qwen2.5-Omni (7B) \cite{xu2025qwen2}. Results for these baselines are derived directly from their official open-source reports. Additionally, we report results for text-only models Qwen3-4B and Qwen3-8B \cite{yang2025qwen3} to serve as theoretical upper bounds for pure language understanding.

\textbf{Results and Analysis.} 
The quantitative results in Table~\ref{tab:openvoice} demonstrate two key findings. 
First, A\textsuperscript{2}-LLM achieves superior performance among audio-native systems, securing the highest scores on \textit{Alpaca Eval} (74.20) and \textit{Trivia QA} (79.90). Notably, for \textit{Reasoning QA}, the linguistic misalignment between our English-only fine-tuning data and the Chinese benchmark leads to occasional instruction non-compliance. Second, and more importantly, our model maintains performance parity with the text-only baselines. This confirms that our training strategy (Sec.~\ref{sec:met:MLLM}) effectively aligns the visual modality without inducing catastrophic forgetting. Rather than competing for model capacity, the generation of facial dynamics proves to be semantically self-consistent with the linguistic backbone, ensuring that expressive output remains grounded in high-level reasoning.

\begin{table}[htbp]
    \centering
    \caption{Evaluation on OpenVoiceBench. Text-only Qwen3 models serve as the semantic upper bound. Best audio-native results are in \textbf{bold}. Abbreviations: Alp.: AlpacaEval, LQ: Llama Questions, RQA: Reasoning QA, TQA: Trivia QA, WQ: Web Questions.}
    \label{tab:openvoice}
    \small
    \setlength{\tabcolsep}{6pt}
    \begin{tabular}{lccccc}
        \toprule
        \textbf{Model} & Alp. & LQ & RQA & TQA & WQ \\
        \midrule
        \multicolumn{6}{c}{\textit{Text-Only (Upper Bound)}} \\
        \midrule
        Qwen3-4B & 76.08 & 75.70 & 39.75 & 82.66 & 75.64 \\
        Qwen3-8B & 78.19 & 82.16 & 39.29 & 86.39 & 83.93 \\
        \midrule
        \multicolumn{6}{c}{\textit{Audio-Native Models}} \\
        \midrule
        GLM-4-Voice & 48.90 & 71.00 & 26.50 & 46.60 & 51.50 \\
        Baichuan-Audio & 69.60 & 74.50 & \textbf{37.90} & 54.20 & 60.30 \\
        StepAudio-Chat & 56.53 & 72.33 & 30.00 & 56.80 & \textbf{73.00} \\
        Qwen2.5-Omni & 72.76 & 75.33 & 31.88 & 57.06 & 62.80 \\
        \textbf{A\textsuperscript{2}-LLM} & \textbf{74.20} & \textbf{75.60} & 34.60 & \textbf{79.90} & 60.60 \\
        \bottomrule
    \end{tabular}
\end{table}

\begin{figure}[t]
    \centering
    \includegraphics[width=1.0\linewidth]{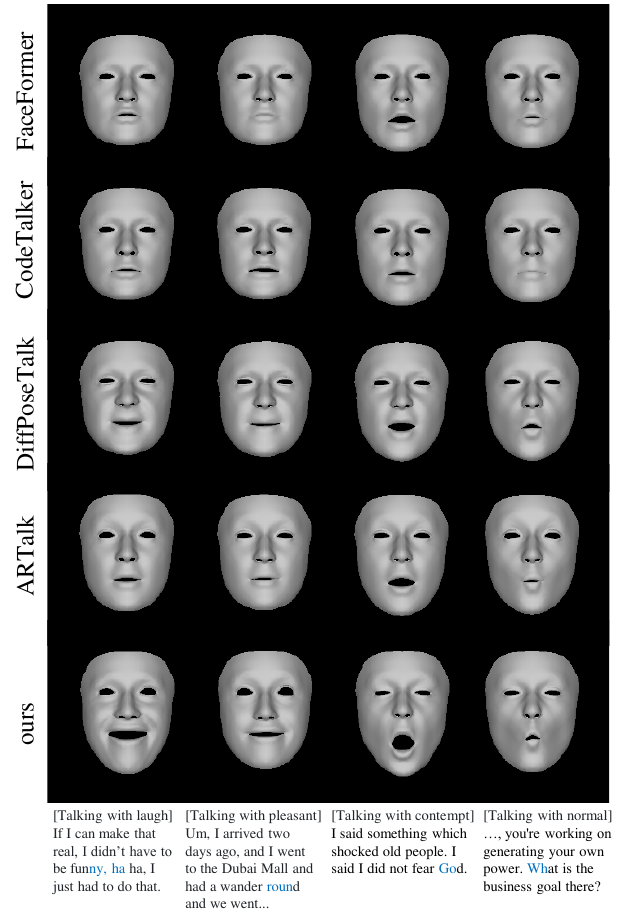} 
    \caption{Visual comparison of facial expressiveness. A\textsuperscript{2}-LLM can spontaneously generate facial motions that better match the content of the reply without requiring explicit emotion conditioning.}
    \label{fig:qualitative}
    \vspace{-5pt}
\end{figure}

\subsection{Facial Motion Quality and Dynamics}
\label{sec:exp:motion_quality}

Evaluating the facial motion quality of an end-to-end conversational model presents a unique challenge: the \textit{Generative Gap}. Unlike traditional audio-driven tasks that reconstruct motion from fixed audio, A\textsuperscript{2}-LLM generates speech with unique prosody and rhythm that naturally diverge from pre-recorded Ground Truth (GT) datasets. To address this, we implement a multi-dimensional evaluation protocol covering both spatial expressiveness and temporal dynamics.

\textbf{Evaluation Protocol: The Pseudo-Oracle Strategy.}
We adopt a Reference Model Comparison strategy using \textbf{DiffPoseTalk}~\cite{sun2024diffposetalk} as a \textit{Pseudo-Oracle}. As a high-fidelity diffusion model prioritizing articulatory precision over real-time efficiency, DiffPoseTalk serves as an ideal "upper bound" for lip synchronization. By generating reference motions from the \textit{same} audio synthesized by A\textsuperscript{2}-LLM, we establish a common ground to compare against leading baselines (ARTalk \cite{chu2025artalk}, FaceFormer \cite{fan2022faceformer}, CodeTalker \cite{xing2023codetalker}). All evaluations are performed on the FLAME-QA held-out test set.

\textbf{Spatial Analysis: Synchronization and Expressiveness.}
We first evaluate frame-level spatial quality using \textbf{Mouth Opening Difference (MOD)} for lip-sync precision and \textbf{Upper Face Dynamics (UFD)} for emotional expressiveness.

\begin{table}[h]
    \centering
    \caption{Spatial quality comparison. A\textsuperscript{2}-LLM achieves competitive lip synchronization (MOD) while demonstrating significantly superior upper-face expressiveness (UFD) compared to audio-driven baselines.}
    \label{tab:expression_metrics}
    \resizebox{0.9\linewidth}{!}{
    \begin{tabular}{l|cc}
        \toprule
        \textbf{Model} & \textbf{MOD} ($\downarrow$) & \textbf{UFD} ($\uparrow$) \\
        \midrule
        ARTalk & \textbf{4.60 $\pm$ 0.80} & 9.40 $\pm$ 1.29 \\
        CodeTalker & 5.29 $\pm$ 0.84 & 2.38 $\pm$ 0.17 \\
        FaceFormer & 5.75 $\pm$ 0.90 & 3.14 $\pm$ 0.35 \\
        \textbf{A\textsuperscript{2}-LLM (Ours)} & 5.08 $\pm$ 0.88 & \textbf{11.13 $\pm$ 1.48} \\
        \bottomrule
    \end{tabular}
    }
\end{table}

As shown in Table~\ref{tab:expression_metrics}, A\textsuperscript{2}-LLM achieves a competitive MOD score (5.08), comparable to the strong baseline ARTalk and outperforming others. This confirms that our end-to-end architecture successfully learns fine-grained articulatory controls. More importantly, A\textsuperscript{2}-LLM dominates in expressiveness (UFD: 11.13). While audio-centric baselines tend to produce a "frozen" upper face due to weak audio-visual correlations, our model leverages semantic understanding to inject context-aware expressions (e.g., surprise-induced eyebrow raises), delivering a truly embodied conversational experience.

\textbf{Temporal Analysis: Frequency and Dynamics.}
Beyond spatial alignment, realistic animation requires physical plausibility and temporal consistency. We analyze motion dynamics in the frequency domain, comparing our model against the strongest baseline, ARTalk.

\begin{table}[h]
    \centering
    \caption{Temporal dynamics analysis. A\textsuperscript{2}-LLM exhibits superior rhythmic consistency (Correlation) and natural motion amplitude (Liveliness) compared to ARTalk.}
    \label{tab:dynamics}
    \resizebox{1.0\linewidth}{!}{
    \begin{tabular}{l|cc}
        \toprule
        \textbf{Metric} & \textbf{ARTalk} & \textbf{A\textsuperscript{2}-LLM (Ours)} \\
        \midrule
        Temporal Correlation ($\uparrow$) & 0.218 & \textbf{0.464} \\
        Velocity Correlation ($\uparrow$) & -0.309 & \textbf{0.111} \\
        Lip Width Correlation ($\uparrow$) & 0.477 & \textbf{0.604} \\
        Liveliness Ratio (Target $\approx$ 1.0) & 0.804 & \textbf{1.087} \\
        Peak Align (ms) ($\downarrow$) & 116.6 & \textbf{114.3} \\
        \bottomrule
    \end{tabular}
    }
\end{table}

The results in Table~\ref{tab:dynamics} demonstrate A\textsuperscript{2}-LLM's superior dynamic fidelity across five key dimensions:
(1) \textbf{Global Rhythm:} \textit{Temporal Correlation}, which measures the synchronization of overall motion trends with the reference, shows a significant improvement (+112\%), indicating our model better captures the global conversational rhythm.
(2) \textbf{Dynamic Trend:} \textit{Velocity Correlation}, representing the consistency of motion direction (first-order derivative), is positive for our model (0.111) but negative for ARTalk, highlighting our superior directional accuracy.
(3) \textbf{Mouth Deformation:} \textit{Lip Width Correlation}, quantifying horizontal mouth stretching critical for smiles and vowels, shows a 26\% gain, validating finer articulatory control.
(4) \textbf{Motion Energy:} \textit{Liveliness Ratio}, the ratio of predicted motion variance to the reference, reaches 1.09 (closer to the target 1.0), confirming that we avoid the "over-smoothing" effect common in regression models.
(5) \textbf{Latency:} \textit{Peak Align}, measuring the time lag of maximum mouth opening, is reduced to 114ms, ensuring tighter audio-visual synchronization.

\textbf{Qualitative Analysis.}
To verify that the high UFD score translates to meaningful expressiveness, we present a visual comparison in Fig.~\ref{fig:qualitative} across diverse semantic contexts. While baseline methods tend to maintain a neutral expression focusing primarily on lip articulation, A\textsuperscript{2}-LLM demonstrates superior semantic-visual alignment. 

For instance, during the laughter segment (``ha ha''), our model naturally generates cheek raising and a smiling mouth shape; when articulating the contemptuous statement (``I did not fear God''), it renders a subtle frown and scornful expression. Similarly, it distinguishes between a relaxed narrative (``wander round'') and a standard business inquiry, adjusting the facial tension accordingly. These results confirm that A\textsuperscript{2}-LLM successfully leverages the backbone's semantic understanding to drive context-aware, emotionally congruent facial behaviors beyond simple audio-lip synchronization.

\begin{table}[htbp]
    \centering
    \caption{\textbf{User Preference Study Results.} The Win Rate represents the percentage of cases where A\textsuperscript{2}-LLM was ranked higher (more expressive) than the competing baseline in pairwise comparisons ($N=60$).}
    \label{tab:user_study}
    \resizebox{0.9\linewidth}{!}{
    \begin{tabular}{l|c|c|c}
        \toprule
        \textbf{Ours vs. Baseline} & \textbf{Win (\%)} & \textbf{Tie (\%)} & \textbf{Loss (\%)} \\
        \midrule
        Ours vs. DiffPoseTalk (Avatar 1) & \textbf{71.7\%} & 10.0\% & 18.3\% \\
        Ours vs. ARTalk (Avatar 2) & \textbf{75.0\%} & 5.0\% & 20.0\% \\
        \bottomrule
    \end{tabular}
    }
\end{table}
\textbf{User Preference Study} To assess perceived emotional quality, we conducted a subjective user study ($N=60$ samples). Participants ranked randomized video triplets comparing our method against DiffPoseTalk and ARTalk based on expressiveness.
As shown in Table~\ref{tab:user_study}, A\textsuperscript{2}-LLM achieves a preference rate of over 71\% against both baselines. These results suggest that human evaluators tend to perceive the facial dynamics generated by our unified framework as more contextually engaging compared to current audio-driven approaches.

\subsection{Ablation on Joint Training Strategy}
\label{sec:exp:lip_ablation}

To validate the impact of our training strategy on synchronization and motion quality, we compared our final model against an \textbf{Adapter-Only} baseline (Frozen LLM).

\begin{table}[h]
    \centering
    \caption{Ablation on lip dynamics. The \textit{Adapter-Only} baseline suffers from severe latency (515ms) and poor correlation, while \textit{Joint Training} ensures precise phase alignment.}
    \label{tab:lip_dynamics}
    \resizebox{1.0\linewidth}{!}{
    \begin{tabular}{l|cc}
        \toprule
        \textbf{Metric} & \textbf{Adapter-Only} & \textbf{Joint Training} \\
        \midrule
        Temporal Correlation ($\uparrow$) & 0.028 & \textbf{0.464} \\
        Velocity Correlation ($\uparrow$) & 0.019 & \textbf{0.111}\\
        Lip Width Correlation ($\uparrow$) & 0.057 & \textbf{0.604}\\
        Liveliness Ratio ($\approx 1.0$) & 0.850 & \textbf{1.087} \\
        Peak Align (ms) ($\downarrow$) & 515.05 & \textbf{114.30}\\
        \bottomrule
    \end{tabular}
    }
\end{table}

As shown in Table~\ref{tab:lip_dynamics}, the Adapter-Only baseline exhibits severe synchronization lag (Peak Align $\sim$515 ms) and negligible motion correlation ($0.028$), indicating that a frozen LLM fails to anticipate precise articulatory timing. In contrast, Joint Training drastically reduces latency to $114$ ms and restores positive velocity correlation ($0.111$). This confirms that fine-tuning the LLM is essential for aligning the phase and rhythm of facial tokens with the speech signal.

\section{Conclusion}
\label{sec:conc}

In this work, we presented \textbf{A\textsuperscript{2}-LLM}, a pioneering end-to-end framework that unifies language processing, audio generation, and facial animation into a single multimodal large model. By moving beyond traditional cascaded pipelines, we effectively bridged the \textit{Semantic-Emotion Gap}, enabling digital avatars to generate facial expressions that are not merely lip-synced but are deeply grounded in conversational context and emotional intent.

Our contributions are threefold: the proposal of a unified architecture that achieves an interaction latency of $\sim$500ms; the construction of \textbf{FLAME-QA}, the first large-scale instruction-tuning dataset for semantically aware 3D facial animation; and a hierarchical motion modeling strategy that ensures both stability and expressiveness. Extensive experiments demonstrate that A\textsuperscript{2}-LLM significantly outperforms existing audio-driven baselines in terms of dynamic diversity and user preference.

\textbf{Limitations and Future Work.}
Despite these advancements, current limitations point to exciting future directions. First, our model is currently optimized for English; expanding to multilingual support is a priority. Second, while we focus on facial dynamics, extending this end-to-end paradigm to full-body gesture generation remains an open challenge. We believe A\textsuperscript{2}-LLM serves as a solid foundation for the next generation of empathetic and truly interactive digital humans.

\nocite{langley00}

\bibliography{example_paper}
\bibliographystyle{icml2026}

\newpage
\appendix
\onecolumn
\section{Data-processing} 
\label{sec:data}

\textbf{FLAME-QA Construction Pipeline}

We start from approximately $800$K VoxCeleb clips and apply the following automatic pipeline:

\begin{itemize}
  \item \emph{Transcription.} We use Whisper v3 Large to obtain text transcripts for each clip.
  \item \emph{Emotion estimation.} A speech emotion recognition module predicts utterance-level emotion labels, which are later used as auxiliary supervision.
  \item \emph{Cleaning and QA generation.} GPT-5.1 is prompted to clean the transcripts, remove noisy or incomplete segments, and generate corresponding questions, yielding QA pairs. Only segments that pass quality filters (e.g., length, intelligibility) are retained, resulting in about $100$K QA pairs.
  \item \emph{Question audio synthesis.} We synthesize spoken questions from the generated text using the IndexTTS2 TTS model, obtaining paired question audio, answer audio (original), and text.
\end{itemize}

\textbf{Emotion-rich synthetic QA.}
For the emotion-rich subset, we first sample and edit QA texts with a large language model to emphasize diverse emotional content and personalized styles. The QA texts are converted to audio via TTS and then fed into InfiniteTalk to generate talking-head videos with rich expressions, yielding roughly $0.8$K clips. SMIRK is applied to all videos to extract frame-level FLAME expression and pose parameters, which serve as ground truth motion supervision.

We use GPT-5.1 as the primary data-cleaning engine. Specifically, we feed the raw ASR transcripts, together with carefully designed prompts (see Fig.~\ref{fig:prompt_clean}), into GPT-5.1 to correct recognition errors, remove disfluencies and irrelevant content, and construct high-quality QA pairs.

\begin{figure}[h]
  \centering
  \begin{tcolorbox}[title={Prompt for Transcript Cleaning and Filtering},
                    colback=gray!3, colframe=gray!50]
    \textbf{System:}\\
    You are a helpful assistant specialized in evaluating text quality
    and generating questions for QA training datasets.

    \medskip
    \textbf{User:}\\
    Evaluate if the following text is suitable as an answer in a QA training dataset
    (complete English sentence, grammatically correct, meaningful content).

    Text: \texttt{\{answer\_text\}}

    If suitable, generate a natural question. If not suitable, return \texttt{null} for the question.

    Respond in JSON format:

    \texttt{\{}\\
    \hspace*{1.5em}\texttt{"is\_suitable": true or false,}\\
    \hspace*{1.5em}\texttt{"question": "your question here" or null}\\
    \texttt{\}}

    Respond ONLY with the JSON object.
  \end{tcolorbox}
  \caption{Prompt used by GPT-5.1 to clean ASR transcripts, assess their suitability as QA answers, and generate corresponding questions in JSON format for constructing the FLAME-QA dataset.}
  \label{fig:prompt_clean}
\end{figure}

\section{Implementation Details}
\label{sec:imp_details}

\textbf{Hardware and Environment.}
All experiments, including the training of the Residual Motion Tokenizer and the instruction tuning of A\textsuperscript{2}-LLM, are conducted on a computational node equipped with $8 \times$ NVIDIA A100 (80GB) GPUs. The framework is implemented using PyTorch.

\textbf{RVQ-VAE Architecture and Training.}
The Residual Motion Tokenization module compresses continuous FLAME parameter sequences ($25$ fps) into discrete codes at a $5$ Hz resolution (temporal group size $G=5$). The architecture comprises an encoder and decoder with a base channel width of $128$ and $3$ residual blocks each. The latent embedding dimension is $256$, discretized by $N_q=6$ hierarchical quantizers, each with a codebook size of $256$.

We optimize the model using AdamW ($\text{lr} = 1 \times 10^{-4}$) for $6$ epochs with a global batch size of $512$ ($64$ per GPU). Regularization includes a dropout rate of $0.05$ and a codebook commitment loss weight of $\gamma=0.25$. The vector quantization loss weight is set to $\lambda_{\text{vq}}=1.0$.

\textbf{Reconstruction Objectives.}
To ensure holistic motion fidelity, we optimize a composite objective. The total reconstruction loss combines parameter-space regression with vertex-space constraints:
\begin{equation}
    \mathcal{L}_{\text{rec}} = \mathcal{L}_{\text{param}} + 10^5 (\mathcal{L}_{\text{lips}} + \mathcal{L}_{\text{face}}) + 10^2 (\mathcal{L}_{\text{vel}} + \mathcal{L}_{\text{acc}})
\end{equation}

The components and their specific configurations are defined as follows:

\textit{1) Parameter Regression:} We enforce a baseline consistency in the 58-dimensional FLAME parameter space with a unit weight ($\lambda_{\text{param}}=1.0$):
\begin{equation}
    \mathcal{L}_{\text{param}} = \| \mathbf{M} - \hat{\mathbf{M}} \|_2^2
\end{equation}

\textit{2) Spatial Geometric Consistency:} To capture fine-grained morphology, we map parameters to 3D vertices in a canonical zero-pose space. To compensate for the small magnitude of metric-scale vertex coordinates, we assign high weights ($\lambda_{\text{lips}} = \lambda_{\text{face}} = 1.0 \times 10^5$) to the $L_2$ distances of the lip and face regions:
\begin{equation}
    \mathcal{L}_{\text{geo}} = \| \mathbf{V}_{lips} - \hat{\mathbf{V}}_{lips} \|_2^2 + \| \mathbf{V}_{face} - \hat{\mathbf{V}}_{face} \|_2^2
\end{equation}

\textit{3) Temporal Dynamics Consistency:} To enforce smoothness and suppress jitter, we constrain the velocity ($\Delta \mathbf{V}$) and acceleration ($\Delta^2 \mathbf{V}$) of the full vertex sequence. These terms are weighted by $\lambda_{\text{vel}} = \lambda_{\text{acc}} = 1.0 \times 10^2$:
\begin{equation}
    \mathcal{L}_{\text{dyn}} = \| \Delta \mathbf{V} - \Delta \hat{\mathbf{V}} \|_2^2 + \| \Delta^2 \mathbf{V} - \Delta^2 \hat{\mathbf{V}} \|_2^2
\end{equation}

\textbf{Motion Connector Configuration.}
The Motion Connector serves as the bridge between the LLM backbone and the discrete motion space. It is instantiated as a lightweight Transformer decoder consisting of $6$ layers. We set the hidden dimension to $768$ and utilize $12$ attention heads per layer to capture complex audio-motion dependencies. To prevent overfitting during the joint alignment phase, a dropout rate of $0.1$ is applied throughout the network.

\textbf{LoRA Configuration.}
To efficiently adapt the LLM backbone for multimodal generation without incurring excessive computational costs, we apply Low-Rank Adaptation (LoRA). We configure the rank to $r=64$ with a scaling factor of $\alpha=32$ and a dropout rate of $0.05$. The adapters are injected into all linear layers of the transformer blocks, specifically targeting the attention projections (\texttt{q\_proj}, \texttt{k\_proj}, \texttt{v\_proj}, \texttt{o\_proj}) and the feed-forward network modules (\texttt{gate\_proj}, \texttt{up\_proj}, \texttt{down\_proj}).

\section{Evaluation Metrics}
\label{sec:metrics}

To comprehensively evaluate the quality of the generated 3D facial animation, we employ a diverse set of metrics covering geometric accuracy, expressive dynamics, and temporal synchronization. All metrics are computed in the canonical zero-pose space (i.e., with global rotation and head translation removed) to focus purely on facial deformations.

\subsection{Spatial and Expressiveness Metrics}

\textbf{Mouth Opening Distance (MOD).}
MOD measures the geometric precision of lip synchronization. It calculates the Mean Absolute Error (MAE) of the vertical mouth opening between the predicted and ground-truth sequences.
Let $h(\mathbf{V})$ denote the vertical mouth opening (distance between upper and lower lips) for a given vertex frame $\mathbf{V}$. The metric is defined as:
\begin{equation}
    \text{MOD} = \frac{1}{T} \sum_{t=1}^{T} | h(\hat{\mathbf{V}}_t) - h(\mathbf{V}_t) | \times 1000
\end{equation}
where $\hat{\mathbf{V}}$ and $\mathbf{V}$ are the predicted and ground-truth vertices, respectively. The result is reported in millimeters (mm). Lower values indicate more accurate lip closure and aperture.

\textbf{Upper Face Dynamics (UFD).}
UFD is a reference-free metric designed to quantify the intensity of upper-face movements (e.g., eye blinks, eyebrow raises). It measures the frame-to-frame change rate of vertex displacements relative to a neutral template.
First, we compute the displacement norm for each vertex $i$ in the upper face region relative to a neutral face $\mathbf{V}_{neu}$: $d_{t,i} = \| \mathbf{v}_{t,i} - \mathbf{v}_{neu,i} \|_2$. The UFD is then calculated as the average temporal change of this displacement:
\begin{equation}
    \text{UFD} = \frac{1}{T-1} \sum_{t=1}^{T-1} \left( \frac{1}{|S_{up}|} \sum_{i \in S_{up}} | d_{t,i} - d_{t-1,i} | \right) \times 10^5
\end{equation}
where $S_{up}$ denotes the set of upper face indices. Higher values indicate richer and more dynamic facial expressions.

\subsection{Temporal Dynamics and Synchronization}

To evaluate the physical plausibility and fine-grained synchronization of the generated motion, we analyze the mouth opening signal $O_t = h(\mathbf{V}_t)$ and the mouth width signal $W_t$ (horizontal distance between lip corners) in the frequency and time domains.

\textbf{1) Temporal Correlation:}
This metric measures the global rhythmic alignment. It computes the Pearson Correlation Coefficient (PCC) between the predicted and ground-truth mouth opening sequences:
\begin{equation}
    \rho_{\text{temp}} = \frac{\text{Cov}(O_{pred}, O_{gt})}{\sigma(O_{pred})\sigma(O_{gt})}
\end{equation}
Higher values indicate that the generated motion follows the correct opening-closing rhythm.

\textbf{2) Velocity Correlation:}
To assess the directional consistency of motion, we compute the correlation of the velocity profiles (first-order derivatives). Let $v_t = O_t - O_{t-1}$. The metric is defined as:
\begin{equation}
    \rho_{\text{vel}} = \text{PCC}(v_{pred}, v_{gt})
\end{equation}
A positive value implies that the avatar opens and closes its mouth in strict phase synchronization with the ground truth.

\textbf{3) Lip Width Correlation:}
While vertical opening dominates speech, horizontal stretching is critical for emotive expressions (e.g., smiles). We compute the Pearson correlation of the mouth width sequences:
\begin{equation}
    \rho_{\text{width}} = \text{PCC}(W_{pred}, W_{gt})
\end{equation}

\textbf{4) Liveliness Ratio:}
Regression-based models often suffer from "over-smoothing," resulting in muted motion. The Liveliness Ratio compares the standard deviation of the velocity (motion energy) between prediction and ground truth:
\begin{equation}
    R_{\text{live}} = \frac{\sigma(v_{pred})}{\sigma(v_{gt}) + \epsilon}
\end{equation}
A value close to $1.0$ indicates that the generated motion possesses natural dynamic amplitude, neither over-smoothed nor jittery.

\textbf{5) Peak Alignment Latency (Peak Align):}
This metric quantifies the synchronization delay. We detect local maxima (peaks) in the mouth opening curves for both sequences. For each peak in the ground truth $p_{gt}^{(i)}$, we find the closest peak in the prediction $p_{pred}^{(j)}$ and calculate the median absolute time difference:
\begin{equation}
    \text{Lag} = \text{Median}\left( \min_{j} | t(p_{gt}^{(i)}) - t(p_{pred}^{(j)}) | \right)
\end{equation}
The result is converted to milliseconds (ms). Lower values indicate tighter synchronization with the speech signal.

\end{document}